\documentclass[10pt,twocolumn,letterpaper]{article}

\pdfoutput=1

\usepackage{iccv}
\usepackage{times}
\usepackage{epsfig}
\usepackage{graphicx}
\usepackage{amsmath}
\usepackage{amssymb}

\usepackage{multirow}
\usepackage{amsthm}

\usepackage[pagebackref=true,breaklinks=true,letterpaper=true,colorlinks,bookmarks=false]{hyperref}

\iccvfinalcopy 

\def\iccvPaperID{2270} 

\ificcvfinal\fi

\begin{document}

\title{Linear Differential Constraints for Photo-polarimetric Height Estimation}

\author{Silvia Tozza\\
Sapienza - Universit{\`a} di Roma\\
{\tt\small tozza@mat.uniroma1.it}
\and
William A. P. Smith\\
University of York\\
{\tt\small william.smith@york.ac.uk}
\and
Dizhong Zhu\\ 
University of York\\
{\tt\small dz761@york.ac.uk}
\and
Ravi Ramamoorthi\\
UC San Diego \\ 
{\tt\small ravir@cs.ucsd.edu}
\and
Edwin R. Hancock\\ 
University of York\\
{\tt\small edwin.hancock@york.ac.uk}
}

\maketitle
\thispagestyle{empty}

\begin{abstract}
In this paper we present a differential approach to photo-polarimetric shape estimation. We propose several alternative differential constraints based on polarisation and photometric shading information and show how to express them in a unified partial differential system. Our method uses the image ratios technique to combine shading and polarisation information in order to directly reconstruct surface height, without first computing surface normal vectors. Moreover, we are able to remove the non-linearities so that the problem reduces to solving a linear differential problem. We also introduce a new method for estimating a polarisation image from multichannel data and, finally, we show it is possible to estimate the illumination directions in a two source setup, extending the method into an uncalibrated scenario. From a numerical point of view, we use a least-squares formulation of the discrete version of the problem. To the best of our knowledge, this is the first work to consider a unified differential approach to solve photo-polarimetric shape estimation directly for height. Numerical results on synthetic and real-world data confirm the effectiveness of our proposed method. 
\end{abstract}

\section{Introduction}\label{sec:intro}

A recent trend in photometric \cite{ecker2010polynomial,MQ16,MF13,TMDDB16,SF16,QMD16} and physics-based \cite{SRT16} shape recovery has been to develop methods that solve directly for surface height, rather than first estimating surface normals and then integrating them into a height map. Such methods are attractive since: 1. they only need solve for a single height value at each pixel (as opposed to the two components of surface orientation), 2. integrability is guaranteed, 3. errors do not accumulate through a two step pipeline of shape estimation and integration and 4. it enables combination with cues that provide depth information directly \cite{kadambi2015polarized}. In both photometric stereo \cite{MQ16,MF13,SF16} and recently in shape-from-polarisation (SfP) \cite{SRT16}, such a direct solution was made possible by deriving equations that are linear in the unknown surface gradient.

In this paper, we explore the combination of SfP constraints with photometric constraints (i.e. photo-polarimetric shape estimation) provided by one or two light sources. Photometric stereo with three or more light sources is a very well studied problem with robust solutions available under a range of different assumptions. Two source photometric stereo is still considered a difficult problem \cite{queau2017photometric} even when the illumination is calibrated and albedo is known. We show that various formulations of one and two source photo-polarimetric stereo lead to the same general problem
(in terms of surface height), 
that illumination can be estimated and that certain combinations of constraints lead to an albedo invariant formulation. Hence, with only modest additional data capture requirements (a polarisation image rather than an intensity image), we arrive at an approach for uncalibrated two source photometric stereo.
We make the following novel contributions:
\begin{itemize}
\vspace{-0.2cm}
\item We show how to estimate a polarisation image from multichannel data such as from colour images, multiple light source data or both (Sec.~\ref{sec:multichanpol}).
\vspace{-0.2cm}
\item We show how polarisation and photometric constraints (Sec.~\ref{Sec:constraints}) can be expressed in a unified formulation (of which previous work \cite{SRT16} is a special case) and that various combinations of these constraints provide different practical advantages (Sec.~\ref{sec:several_combinations}).
\vspace{-0.2cm}
\item We show how to estimate the illumination directions in two source photo-polarimetric data leading to an uncalibrated solution (Sec.~\ref{sect:light_estimation}).
\end{itemize}

\subsection{Related Work}

The polarisation state of light reflected by a surface provides a cue to the material properties of the surface and, via a relationship with surface orientation, the shape. Polarisation has been used for a number of applications, including early work on material segmentation \cite{wolff1991constraining} and diffuse/specular reflectance separation \cite{nayar1997separation}. However, there has been a resurgent interest \cite{SRT16,kadambi2015polarized,taamazyan2016shape,ngo2015shape} in using polarisation information for shape estimation.

\noindent {\bf Shape-from-polarisation}\ \  The degree to which light is linearly polarised and the orientation associated with maximum reflection are related to the two degrees of freedom of surface orientation. In theory, this polarisation information alone restricts the surface normal at each pixel to two possible directions. Both Atkinson and Hancock \cite{AH2006} and Miyazaki \etal \cite{miyazaki2003polarization} solve the problem of disambiguating these polarisation normals via propagation from the boundary under an assumption of global convexity. Huynh \etal \cite{huynh2010shape} also disambiguate polarisation normals with a global convexity assumption but estimate refractive index in addition. These works all 
used a diffuse polarisation model while Morel \etal \cite{morel2005polarization} use a specular polarisation model for metals. Recently, Taamazyan \etal \cite{taamazyan2016shape} introduced a mixed specular/diffuse polarisation model. All of these methods estimate surface normals that must be integrated into a height map. Moreover, since they rely entirely on the weak shape cue provided by polarisation and do not enforce integrability, the results are extremely sensitive to noise.

\noindent {\bf Photo-polarimetric methods}\ \  There have been a number of attempts to combine photometric constraints with polarisation cues. Mahmoud \etal \cite{mahmoud2012direct} used a shape-from-shading cue with assumptions of known light source direction, known albedo and Lambertian reflectance to disambiguate the polarisation normals. Atkinson and Hancock \cite{atkinson2007surface} used calibrated, three source Lambertian photometric stereo for disambiguation but avoiding an assumption of known albedo. Smith \etal \cite{SRT16} showed how to express polarisation and shading constraints directly in terms of surface height, leading to a robust and efficient linear least squares solution. They also show how to estimate the illumination, up to a binary ambiguity, making the method uncalibrated. However, they require known or uniform albedo. We explore variants of this method by introducing additional constraints that arise when a second light source is introduced, allowing us to relax the uniform albedo assumption. 
We also give an explanation for why the matrix they consider is full-rank except in a unique case. 
Recently, Ngo \etal \cite{ngo2015shape} derived constraints that allowed surface normals, light directions and refractive index to be estimated from polarisation images under varying lighting. However, this approach requires at least 4 lights. All of the above methods operate on single channel images and do not exploit the information available in colour images.

\noindent {\bf Polarisation with additional cues}\ \  Rahmann and Canterakis \cite{rahmann2001reconstruction} combined a specular polarisation model with stereo cues. Similarly, Atkinson and Hancock \cite{atkinson2007shape} used polarisation normals to segment an object into patches, simplifying stereo matching. Stereo polarisation cues have also been used for transparent surface modelling \cite{miyazaki2004transparent}. Huynh \etal \cite{huynh2013shape} extended their earlier work to use multispectral measurements to estimate both shape and refractive index. Drbohlav and Sara \cite{drbohlav2001unambiguous} showed how the Bas-relief ambiguity \cite{belhumeur1999bas} in uncalibrated photometric stereo could be resolved using polarisation. However, this approach requires a polarised light source. 
Recently, Kadambi \etal \cite{kadambi2015polarized} proposed an interesting approach in which a single polarisation image is combined with a depth map obtained by an RGBD camera. The depth map is used to disambiguate the normals and provide a base surface for integration.

\section{Representing Polarisation Information}\label{Sec:assumptions}
We place a camera at the origin of a three-dimensional coordinate system (\emph{Oxyz}) in such a way that \emph{Oxy} coincides with the image plane and \emph{Oz} with the optical axis. In Sec.~\ref{sec:several_combinations} we propose a unified formulation for a variety of methods, all of which assume a) orthographic projection, b) known refractive index of the surface.
Other assumptions will be given later on, depending on the specific problem at hand. 
We  denote by ${\mathbf v}$ the  viewer direction, by ${\mathbf s}$ a general light source direction with ${\mathbf v} \neq {\mathbf s}$. We only require the third components of these unit vectors to be greater than zero (\ie all the vectors belong to the upper hemisphere). We will denote by ${\mathbf t}$ a second light source where required. 
We parametrise the unknown surface  height by the function $z({\mathbf x})$, where ${\mathbf x}=(x,y)$ is an image location, and 
the unit normal to the 
surface at the point ${\mathbf x}$ is given by:
\begin{equation} \label{normal_N}
{\mathbf n}({\mathbf x}) = \frac{\hat{\mathbf n}({\mathbf x})}{|\hat{\mathbf n}({\mathbf x})|} = \frac{[-z_x, -z_y, 1]^T}{\sqrt{1 + |\nabla z({\mathbf x})|^2}}, 
\end{equation} 
where $\hat{\mathbf n}({\mathbf x})$ is the outgoing normal vector and $z_x, z_y$ denotes the partial derivative of $z({\mathbf x})$ w.r.t. x and y, respectively, so that $\nabla z({\mathbf x}) = (z_x, z_y)$. 
We now  introduce relevant polarization theory, describing how we can estimate a polarisation image from multichannel data. 

\subsection{Polarisation image}

When unpolarised light is reflected by a surface it becomes partially polarised~\cite{wolff1997polarization}.
A {\it polarisation image} can be estimated by capturing a sequence of images  in which a linear polarising filter in front of the camera lens  is rotated through a sequence of $P\geq 3$ different angles $\vartheta_j$, $j\in \left\{1,\ \dots,\ P\right\}$. The measured intensity at a pixel varies sinusoidally with the polariser angle:
\begin{equation}\label{eqn:TRS}
 i_{\vartheta_j}({\mathbf x})=i_{\textrm{un}}({\mathbf x}) \, \big( 1 + \rho({\mathbf x}) \cos(2\vartheta_j-2\phi({\mathbf x}))\big). 
\end{equation}
The polarisation image is thus obtained by decomposing the sinusoid at every pixel location  into three quantities~\cite{wolff1997polarization}: the {\it phase angle}, $\phi({\mathbf x})$, the {\it degree of polarisation}, $\rho({\mathbf x})$, and the {\it unpolarised intensity}, $i_{\textrm{un}}({\mathbf x})$.  
The parameters of the sinusoid can be estimated from the captured image sequence using non-linear least squares~\cite{AH2006}, linear methods~\cite{huynh2010shape} or via a closed form solution~\cite{wolff1997polarization} for the specific case of $P=3$, $\vartheta\in\{ 0^{\circ}, 45^{\circ}, 90^{\circ}\}$. 

\subsection{Multichannel polarisation image estimation}
\label{sec:multichanpol}

A polarisation image is usually computed by fitting the sinusoid in \eqref{eqn:TRS} to observed data in a least squares sense. Hence, from $P\geq 3$ measurements we estimate $i_{\textrm{un}}$, $\rho$ and $\phi$. In practice, we may have access to multichannel measurements. For example, we may capture colour images (3 channels), polarisation images with two different light source directions (2 channels) or both (6 channels). Since $\rho$ and $\phi$ depend only on surface geometry (assuming that, in the case of colour images, the refractive index does not vary with wavelength), then we expect these quantities to be constant over the channels. On the other hand, $i_{\textrm{un}}$ will vary between channels either because of a shading change caused by the different lighting or because the albedo or light source intensity is different in the different colour channels. Hence, in a multichannel setting with $C$ channels, we have $C+2$ unknowns and $CP$ observations. If we use information across all channels simultaneously, the system is more constrained and the solution  will be more robust to noise. Moreover, we do not need to make an arbitrary choice about the channel from which we estimate the polarisation image. This idea shares something in common with that of Narasimhan \etal \cite{Narasimhan_2003_5038}, though their material/shape separation was not in the context of polarisation.

Specifically, we can express the multichannel observations in channel $c$ with polariser angle $\vartheta_j$ as
\begin{equation}
    i^c_{\vartheta_j}({\mathbf x}) =  
    i^c_{\textrm{un}}({\mathbf x})(1 + \rho({\mathbf x})\cos(2\vartheta_j-2\phi({\mathbf x}))).
\end{equation}
The system of equations is linear in the unpolarised intensities and, by a change of variables, can be made linear in $\rho$ and $\phi$ \cite{huynh2010shape}. Hence, we wish to solve a bilinear system and do so in a least squares sense using interleaved alternating minimisation. Specifically, we a) fix $\rho$ and $\phi$ and then solve linearly for the unpolarised intensity in each channel and b) then fix the unpolarised intensities and solve linearly for $\rho$ and $\phi$ using  all channels simultaneously. Concretely, for a single  pixel, we obtain the unpolarised intensities across channels by solving:
\begin{equation}
\min_{i^1_{\textrm{un}}({\mathbf x}), \dots, i^C_{\textrm{un}}({\mathbf x})} \left\| {\bf C}_I \left[
  i^1_{\textrm{un}}({\mathbf x}),
  \dots,
  i^C_{\textrm{un}}({\mathbf x})
\right]^T - {\bf d}_I
\right\|^2 ,
\end{equation}
where ${\bf C}_I\in\mathbb{R}^{CP\times C}$ is given by
{\small{
\begin{equation}
{\bf C}_I =
\begin{bmatrix}
 (1+ \rho({\mathbf x})\cos(2\vartheta_1-2\phi({\mathbf x}))){\bf I}_C \\
  \vdots \\
( 1+ \rho({\mathbf x})\cos(2\vartheta_P-2\phi({\mathbf x}))){\bf I}_C
\end{bmatrix},
\end{equation}
}}
with ${\bf I}_C$  denoting the $C\times C$ identity matrix, 
and ${\bf d}_I\in\mathbb{R}^{CP}$ is given by
{\small{
\begin{equation*}
{\bf d}_I=
\left[
  i^1_{\vartheta_1}({\mathbf x}),
  \dots,
  i^C_{\vartheta_1}({\mathbf x}),
  i^1_{\vartheta_2}({\mathbf x}),
  \dots,
  i^C_{\vartheta_P}({\mathbf x})
\right]^T.
\end{equation*}
}}
Then, with  the unpolarised intensities fixed, we solve for $\rho$ and $\phi$ using the following linearisation:
\begin{equation}
\min_{a,b} \left\|
{\bf C}_{\rho\phi}
\begin{bmatrix}
  a \\
  b
\end{bmatrix}-
{\bf d}_{\rho\phi}\right\|^2,
\end{equation}
where
$[a \,\, b]^T = [\rho({\mathbf x})\cos(2\phi({\mathbf x})), \rho({\mathbf x})\sin(2\phi({\mathbf x}))]^T,
$ and
${\bf C}_{\rho\phi}\in\mathbb{R}^{CP\times 2}$ is given by
{\small{
\begin{equation}
{\bf C}_{\rho\phi}=
    \begin{bmatrix}
 i^1_{\textrm{un}}({\mathbf x})\cos(2\vartheta_1) & i^1_{\textrm{un}}({\mathbf x})\sin(2\vartheta_1) \\
\vdots & \vdots \\
i^1_{\textrm{un}}({\mathbf x})\cos(2\vartheta_P) & i^1_{\textrm{un}}({\mathbf x})\sin(2\vartheta_P) \\
i^2_{\textrm{un}}({\mathbf x})\cos(2\vartheta_1) & i^2_{\textrm{un}}({\mathbf x})\sin(2\vartheta_1) \\
\vdots & \vdots \\
i^C_{\textrm{un}}({\mathbf x})\cos(2\vartheta_P) & i^C_{\textrm{un}}({\mathbf x})\sin(2\vartheta_P) \\
\end{bmatrix},
\end{equation}
}}
and ${\bf d}_{\rho\phi}\in\mathbb{R}^{CP}$ is given by:
{\small{
\begin{equation}{\bf d}_{\rho\phi}=
    \begin{bmatrix}
i^1_{\vartheta_1}({\mathbf x})-i^1_{\textrm{un}}({\mathbf x}) \\
\vdots \\
i^1_{\vartheta_P}({\mathbf x})-i^1_{\textrm{un}}({\mathbf x}) \\
i^2_{\vartheta_1}({\mathbf x})-i^2_{\textrm{un}}({\mathbf x}) \\
\vdots \\
i^C_{\vartheta_P}({\mathbf x})-i^C_{\textrm{un}}({\mathbf x})
\end{bmatrix}.
\end{equation}
}}
We estimate  $\rho$ and $\phi$ from the linear parameters using $\phi({\mathbf x})=\frac{1}{2}\textrm{atan2}(b,a)$ and $\rho({\mathbf x})=\sqrt{a^2+b^2}$. 

\begin{figure*}
    \centering
{\setlength{\tabcolsep}{5pt}
    \begin{tabular}{c|cc|cc}
         \multicolumn{1}{c}{Input} & \multicolumn{2}{c}{Single channel estimation} & \multicolumn{2}{c}{Multichannel estimation} \\
         \includegraphics[height=2cm, clip=true]{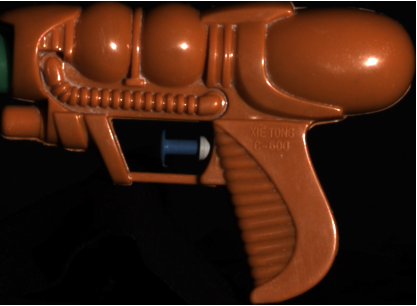}  & 
         \includegraphics[height=2.1cm, trim=192px 352px 177px 347px, clip=true]{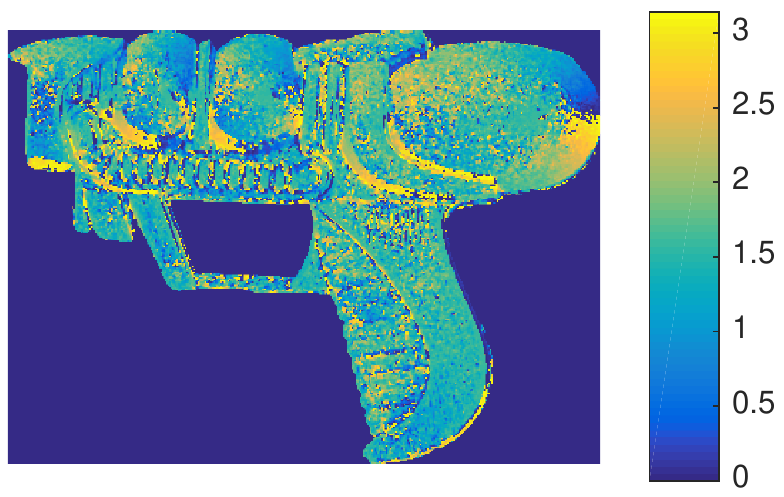} & 
    \includegraphics[height=2.1cm, trim=192px 352px 177px 347px, clip=true]{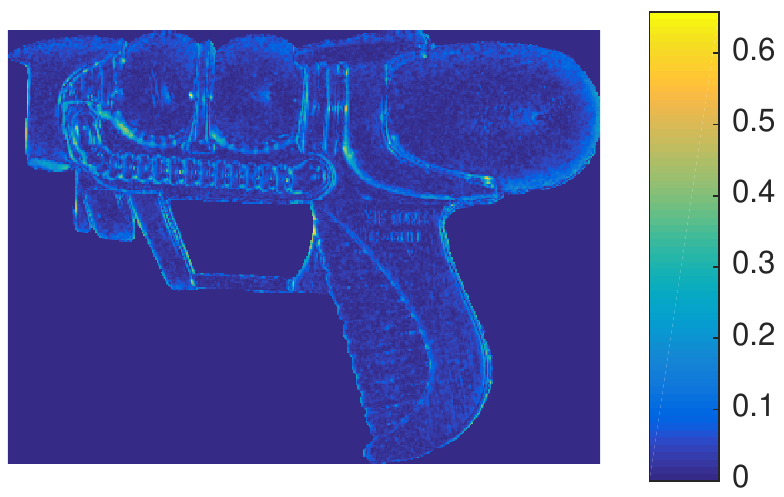} &
        \includegraphics[height=2.1cm, trim=192px 352px 177px 347px, clip=true]{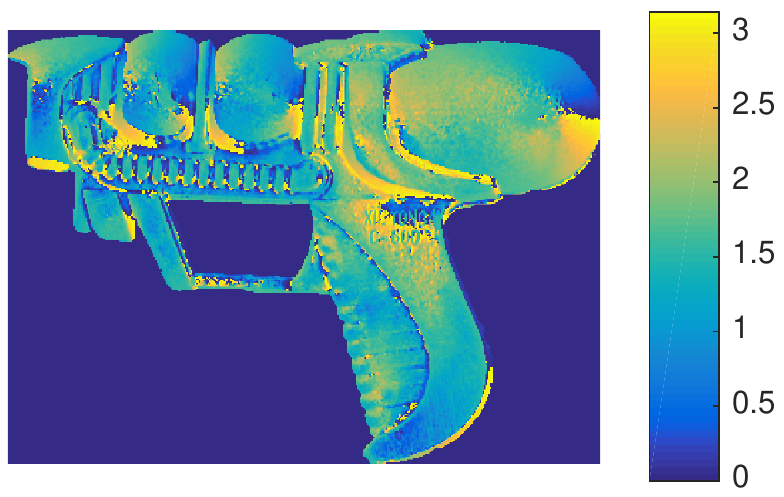} &
    \includegraphics[height=2.1cm, trim=192px 352px 177px 347px, clip=true]{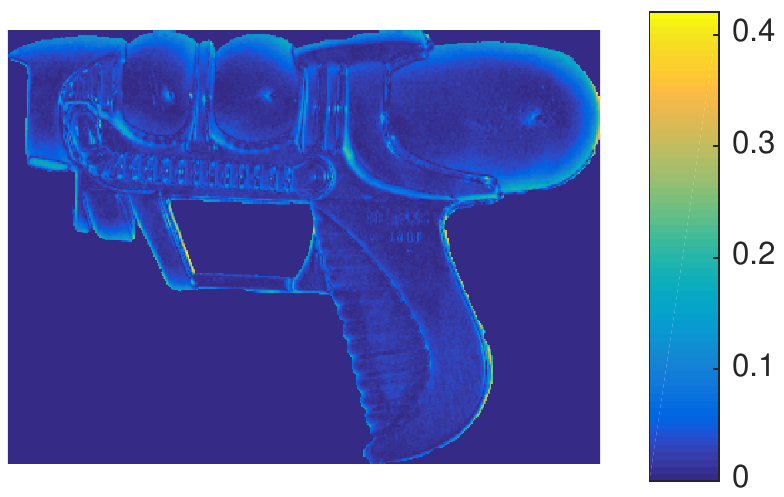}
    \end{tabular}
    }
    \caption{
    Multichannel polarisation image estimation. 
    Left to right: an image from the input sequence; phase angle ($\phi$) and degree of polarisation ($\rho$) estimated from a single channel; phase angle ($\phi$) and degree of polarisation ($\rho$) estimated from three colour channels and two light source directions.}
    \label{fig:pol_image}
\end{figure*}

We initialise by computing a polarisation image from one channel using linear least squares, as in \cite{huynh2010shape}, and then use the estimated $\rho$ and $\phi$ to begin alternating  interleaved optimisation by solving for the unpolarised intensities across channels.
We  interleave and alternate the two steps until convergence. In practice, we find that this approach not only dramatically reduces noise in the polarisation images but also removes the ad hoc step of choosing an arbitrary channel to process. We show an example of the results obtained in Figure \ref{fig:pol_image}. The multichannel result is visibly less noisy than the single channel performance.

\section{Photo-polarimetric height constraints} \label{Sec:constraints}

In this section we describe the different constraints provided by photo-polarimetric information and then  show how to combine them to arrive at linear equations in the unknown  surface height. 
\subsection{Degree of polarisation constraint}\label{subsec:DOP_constr}
A polarisation image provides a constraint on the surface normal direction at each pixel. 
The exact nature of the constraint depends on the polarisation model used.
In this paper we will consider diffuse polarisation, due to subsurface scattering (see~\cite{AH2006} for more details). 
The degree of diffuse polarisation $\rho_d({\mathbf x})$ at each point ${\mathbf x}$ can be expressed in terms of the refractive index $\eta$ and the surface zenith angle $\theta \in [0, \frac{\pi}{2}]$ as follows (Cf.~\cite{AH2006}):
\begin{eqnarray} \label{eq:diffuse_degree_pol}
&&\rho_d({\mathbf x}) =  \\
&&\hspace{-0.8cm} \frac{(\eta-1/\eta)^2 \sin^2(\theta)}{2\! +\! 2\eta^2 \!-\! (\eta \!+\! 1/\eta)^2 \sin^2(\theta) \! +\! 4 \cos(\theta) \sqrt{\eta^2 \!- \sin^2 (\theta)}}. \nonumber
\end{eqnarray}
Recall that the zenith angle is the angle between the unit surface normal vector ${\mathbf n}({\mathbf x})$ and the viewing direction ${\mathbf v}$. If we 
 know the degree of polarisation $\rho_d({\mathbf x})$ and the refractive index $\eta$ (or have good  estimates of  them at hand), 
equation \eqref{eq:diffuse_degree_pol} can be rewritten  
with respect to the cosine of the zenith angle, and expressed in terms of the function, $f(\rho_d({\mathbf x}),\eta)$, that depends on the measured 
degree of polarisation and the refractive index:
{\small{
\begin{align}
&\hspace{0.cm} \cos(\theta) = {\mathbf n}({\mathbf x})\cdot{\mathbf v} = f(\rho_d({\mathbf x}),\eta) =  \label{eq:polariz_constraint} \\
&\sqrt{\frac{\eta^4 \!(1\!-\!\rho_d^2) \!+\! 2\eta^2 (2\rho_d^2 \!+\!\rho_d \!-\!1)\! +\! \rho_d^2 \!+\! 2\rho_d \!-\!4 \eta^3 \rho_d \sqrt{1\!-\!\rho_d^2} \!+\!1}
{(\rho_d+1)^2 \, (\eta^4 + 1) + 2\eta^2(3\rho_d^2 + 2\rho_d -1)}}  \nonumber
\end{align}
}}
where we drop the dependency of $\rho_d$ on $({\mathbf x})$ for brevity.

\subsection{Shading constraint}\label{subsec:shading_constr}
The unpolarised intensity provides an additional constraint on the surface normal direction via an appropriate reflectance model. We assume that pixels have been labelled as diffuse or specular dominant and restrict consideration to diffuse shading. In practice, we deal with specular pixels in the same way as \cite{SRT16} and simply assume that they point in the direction of the halfway vector between $\mathbf{s}$ and $\mathbf{v}$. For the diffuse pixels, we therefore assume that light is reflected according to the Lambert's law. 
Hence, the unpolarised intensity is related to the surface normal by:
\begin{equation}\label{eq:lamb}
i_{\textrm{un}}({\mathbf x}) = \gamma(\mathbf x) \cos(\theta_i) = \gamma(\mathbf x) {\mathbf n}({\mathbf x}) \cdot {\mathbf s},
\end{equation}
 where 
 $\gamma(\mathbf x)$ is the albedo.  
Writing ${\mathbf n}({\mathbf x})$ in terms of the gradient of $z$ as reported in \eqref{normal_N}, \eqref{eq:lamb} can be rewritten as follows:
\begin{equation}\label{eq:lamb2}
i_{\textrm{un}}({\mathbf x}) = \gamma(\mathbf x) \frac{-\nabla z({\mathbf x})\cdot \tilde{{\mathbf s}} + s_3}{\sqrt{1 + |\nabla z({\mathbf x})|^2}},
\end{equation}
with $\tilde{{\mathbf s}} = (s_1, s_2)$. 
This is a non-linear equation, but we will see in Sec.~\ref{Sect:DOP+shading_constr} and \ref{Sect:Shading+shading_constr} how it is possible to remove the non-linearity by using the ratios technique.

\subsection{Phase angle constraint}\label{subsec:phase_angle}
An additional  constraint comes from the phase angle, which determines the azimuth angle of the surface normal $\alpha({\mathbf x}) \in [0, 2\pi]$ up to a $180^{\circ}$ ambiguity. This constraint can be rewritten as a collinearity condition \cite{SRT16}, that is satisfied by either of the two possible azimuth angles implied by the phase angle measurement. 
Specifically, for diffuse pixels we require the projection of the surface normal into the $x$-$y$ plane, $[n_x\ n_y]$, and a vector in the image plane pointing in the phase angle direction, $[\sin(\phi)\ \cos(\phi)]$, to be collinear. This corresponds to requiring 
\begin{equation} \label{eqn:phase}
 \mathbf{n}({\mathbf x})\cdot [\cos(\phi({\mathbf x}))\ -\sin(\phi({\mathbf x}))\ 0]^T = 0.
\end{equation}
In terms of the surface gradient, using \eqref{normal_N}, it is equivalent to 
\begin{equation}\label{phase_constraint}
(-\cos{\phi},\sin{\phi}) \cdot \nabla z = 0.
\end{equation}
A similar expression can be obtained for specular pixels, substituting in the $\frac{\pi}{2}$-shifted phase angles.
The advantage of doing this will become  clear in Sec.~\ref{sec:albedo_inv}.

\subsection{Degree of polarisation ratio constraint}\label{Sect:DOP+shading_constr}
Combining the two constraints illustrated in Sec.~\ref{subsec:DOP_constr} and \ref{subsec:shading_constr}, we can arrive at a linear equation, that we refer to as the DOP ratio constraint. 
Recall that $\cos(\theta) = \mathbf{n}(\mathbf{x}) \cdot \mathbf{v}$ and that we can express $\mathbf{n}$ in terms of the gradient of $z$ by using \eqref{normal_N}, then isolating the non-linear term in \eqref{eq:polariz_constraint} we obtain
\begin{equation}\label{eq1}
\sqrt{1 + |\nabla z({\mathbf x})|^2} = \frac{-\nabla z({\mathbf x})\cdot \tilde{{\mathbf v}} + v_3}{f(\rho_d({\mathbf x}),\eta)},
\end{equation}
where  $\tilde{{\mathbf v}} = (v_1, v_2)$. 
On the other hand, considering the shading information contained in \eqref{eq:lamb2}, and again isolating the non-linearity  
we arrive at the following
\begin{equation}\label{eq2}
\sqrt{1 + |\nabla z({\mathbf x})|^2} = \gamma({\mathbf x}) \frac{-\nabla z({\mathbf x})\cdot \tilde{{\mathbf s}} + s_3}{i_{\textrm{un}}({\mathbf x})}.
\end{equation}
Note that we are supposing ${\mathbf s} \ne {\mathbf v}$, 
and $i_{\textrm{un}}({\mathbf x})\neq 0$, $f(\rho_d({\mathbf x}),\eta) \neq 0$. 
Inspecting  Eqs. \eqref{eq1} and \eqref{eq2} we obtain
\begin{equation}\label{eq:DOP+shad_constr}
\frac{-\nabla z({\mathbf x})\cdot \tilde{{\mathbf v}} + v_3}{f(\rho_d({\mathbf x}),\eta)} = \gamma({\mathbf x}) \frac{-\nabla z({\mathbf x})\cdot \tilde{{\mathbf s}} + s_3}{i_{\textrm{un}}({\mathbf x})}.
\end{equation}
We thus  arrive at  the following partial differential equation (PDE):
\begin{equation}\label{PDE_pb}
{\mathbf b}({\mathbf x})\cdot \nabla z({\mathbf x})=h({\mathbf x}), 
\end{equation}
where
\begin{equation}\label{b_def_DOP+shad}
{\mathbf b}({\mathbf x}) := 
{\mathbf b}^{(f,i_{\textrm{un}})} = 
i_{\textrm{un}}({\mathbf x}) \tilde{{\mathbf v}} - \gamma({\mathbf x})
f(\rho_d({\mathbf x}),\eta) \, \tilde{{\mathbf s}},
\end{equation}
and
\begin{equation}\label{h_def_DOP+shad}
h({\mathbf x}) := 
{h}^{(f,i_{\textrm{un}})}=
i_{\textrm{un}}({\mathbf x})v_3 - \gamma({\mathbf x})
f(\rho_d({\mathbf x}),\eta) \, s_3.
\end{equation} 

\subsection{Intensity ratio constraint}\label{Sect:Shading+shading_constr}

Finally, we construct an intensity ratio constraint by considering two unpolarised images, $i_{\textrm{un},1},i_{\textrm{un},2}$, taken from two different light source directions, $\mathbf{s},\mathbf{t}$. 
We construct our constraint equation by applying \eqref{eq:lamb} twice, once for each light source. We can remove the non-linearity as before and take a ratio, arriving at the following equation:
\begin{equation}\label{eq:shad+shad_constr}
i_{\textrm{un},2}(-\nabla z({\mathbf x})\cdot \tilde{{\mathbf s}} +s_3) = i_{\textrm{un},1}(-\nabla z({\mathbf x})\cdot \tilde{{\mathbf t}} + t_3).
\end{equation}
The above  equation is  independent of albedo, light source intensity and non-linear normalisation term.
Again as before, we can rewrite \eqref{eq:shad+shad_constr} as a PDE in the form  of \eqref{PDE_pb} with
\begin{equation}\label{b_def_shad+shad}
{\mathbf b}({\mathbf x}) :=
{\mathbf b}^{(i_{\textrm{un},1},i_{\textrm{un},2})} = 
i_{\textrm{un},2}({\mathbf x}) \tilde{{\mathbf s}} - i_{\textrm{un},1}({\mathbf x}) \, \tilde{{\mathbf t}},
\end{equation}
where $\tilde{{\mathbf t}} = (t_1, t_2)$,  
and
\begin{equation}\label{h_def_shad+shad}
h({\mathbf x}) := 
h^{(i_{\textrm{un},1},i_{\textrm{un},2})} = 
i_{\textrm{un},2}({\mathbf x})s_3 -
i_{\textrm{un},1}({\mathbf x}) \, t_3.
\end{equation} 

\section{A unified PDE formulation}\label{sec:several_combinations}

Commencing  from the constraints introduced in Sec.~\ref{Sec:constraints}, in this section we show how to solve  several different  problems in photo-polarimetric shape estimation. The common feature is that these are all linear in the unknown height, and are expressed in a unified formulation in terms of 
a system of PDEs in the same general form: 
\begin{equation}\label{PDE_pb_matrix}
{\mathbf B}({\mathbf x}) \nabla z({\mathbf x})={\mathbf h}({\mathbf x}),
\end{equation}
where ${\mathbf B}: \bar{\Omega} \rightarrow {\mathbb R}^{J\times 2}$, ${\mathbf h}: \bar{\Omega} \rightarrow {\mathbb R}^{J\times 1}$, denoting by $\Omega$ the reconstruction domain and being $J=2,3$ or $4$ depending on the cases. 
\eqref{PDE_pb_matrix} 
is a compact and general equation, suitable for  describing several cases in a unified differential formulation that  solves directly for surface  height. 

Different combinations of the three constraints described in Sec.~\ref{Sec:constraints} that are linear in the surface gradient can be combined in the formulation of \eqref{PDE_pb_matrix}. Each corresponds to different assumptions and have different pros and cons. We explore three variants and show that \cite{SRT16} is a special case of our formulation. We summarise the alternative formulations in Tab.~\ref{tab:methods}. 

\begin{table}
    \centering
    \begin{tabular}{|c||c|c|c|}
    \hline
          & Phase & DOP   & Intensity  \\
         Method  & angle & ratio & ratio \\
                \hline
         \cite{SRT16} & \checkmark & \checkmark & \\
         Proposed 1 & \checkmark & & \checkmark \\
         Proposed 2 & & \checkmark & \checkmark \\
         Proposed 3 & \checkmark & \checkmark & \checkmark \\
         \hline
    \end{tabular}
    \caption{Summary of the different formulations}
    \label{tab:methods}
\end{table}

\subsection{Single light and polarisation formulation}

This case has been studied in \cite{SRT16}. It uses a single polarisation image, requires known illumination (though \cite{SRT16} show how this can be estimated if unknown) and assumes that albedo is known or uniform. This last assumption is quite restrictive, since it can only be applied to objects with  homogeneous surfaces. With just a single illumination condition, only the phase angle and DOP ratio constraints are available.
This  thus becomes a special case of our general unified formulation \eqref{PDE_pb_matrix}, 
where $\mathbf{B}$ and ${\mathbf h}$ are defined as
\begin{equation}\label{b_def_eccv_case}
{\mathbf B} =  
\begin{bmatrix}
  {b}^{(f,i_{\textrm{un}})}_1 & {b}^{(f,i_{\textrm{un}})}_2 \\
 -\cos{\phi}  & \sin{\phi} \\
\end{bmatrix},
\quad
\mathbf h = [{h}^{(f,i_{\textrm{un}})}, 0]^T,
\end{equation}
with ${\mathbf b}^{(f,i_{\textrm{un}})}$ and ${h}^{(f,i_{\textrm{un}})}$ defined by \eqref{b_def_DOP+shad} and \eqref{h_def_DOP+shad}, with uniform $\gamma(\mathbf x)$ and ${\mathbf v} = [0,0,1]^T$. 

\subsection{Proposed 1: Albedo invariant formulation}\label{sec:albedo_inv}

Our first proposed method uses
the phase angle constraint \eqref{phase_constraint} and two unpolarised images, taken from two different light source directions, 
obtained through \eqref{eq:lamb2} and combined as in \eqref{eq:shad+shad_constr}. 
In this case the problem studied  is described by the system of PDEs \eqref{PDE_pb_matrix} with
{\small{
\begin{equation}\label{b_def_albedo_ind}
{\mathbf B}({\mathbf x}) = 
\begin{bmatrix}
  {b}^{(i_{\textrm{un},1},i_{\textrm{un},2})}_1 & {b}^{(i_{\textrm{un},1},i_{\textrm{un},2})}_2 \\
 -\cos{\phi}  & \sin{\phi} \\
\end{bmatrix}, {\mathbf h}({\mathbf x})= \begin{bmatrix}
  h^{(i_{\textrm{un},1},i_{\textrm{un},2})} \\
  0
\end{bmatrix},
\end{equation}
}}
where ${\mathbf b}^{(i_{\textrm{un},1},i_{\textrm{un},2})}$ and ${h}^{(i_{\textrm{un},1},i_{\textrm{un},2})}$ defined as in \eqref{b_def_shad+shad} and \eqref{h_def_shad+shad}. 
The phase angle does not depend on albedo and the intensity ratio constraint is invariant to albedo. As a result, this formulation is particularly powerful because it allows albedo invariant height estimation. Moreover, the light source directions in the two images can be estimated (again, in an albedo invariant manner) using the method in Sec.~\ref{sect:light_estimation}. 

Once surface height has been estimated, we can compute the surface normal at each pixel and it is then straightforward to estimate an albedo map using \eqref{eq:lamb}. Where we have two diffuse observations, we can compute albedo from two equations of the form of \eqref{eq:lamb} in a least squares sense. In real data, where we have specular pixel labels, we use only the diffuse observations at each pixel. To avoid artifacts at the boundary of specular regions, we introduce a gradient consistency term into the albedo estimation. We encourage the gradient of the albedo map to match the gradients of the intensity image for diffuse pixels.

\subsection{Proposed 2: Phase invariant formulation}

Our second proposed method uses only the  DOP ratio and the intensity ratio constraints. This means that phase angle estimates are not used. The advantage of this is that phase angles are subject to a shift of $\frac{\pi}{2}$ at specular reflections when compared to diffuse reflections. So, the phase angle constraint relies upon having accurate per-pixel specularity labels, which  classify reflections as either dominantly  specular  or diffuse  (or  alternatively use a mixed polarisation model \cite{taamazyan2016shape} with a four way ambiguity). 
In this case we need a) two unpolarised intensity images, taken with two different 
light source directions, ${\mathbf s}$ and ${\mathbf t}$, obtained through \eqref{eq:lamb2}, b)   polarisation information  from the function $f(\rho,\eta)$ and c) knowledge of the albedo map. We need ${\mathbf s}, {\mathbf t}, {\mathbf v}$ non-coplanar in order to have the matrix field ${\mathbf B}$ not singular. 
Note that the function $f$, obtained from polarization information (as in \eqref{eq:polariz_constraint}), is the same for the two required   images. The reason for this is that it  does not depend on the light source directions but only on the viewer direction  ${\mathbf v}$  which does not change. 
This formulation can be deduced starting from \eqref{eq:shad+shad_constr} and \eqref{eq:DOP+shad_constr},
arriving at a PDE system as in \eqref{PDE_pb_matrix} with
\begin{equation}\label{b_def_phase_ind}
{\mathbf B} =[{\mathbf b}^{(f,i_{\textrm{un},1})} , {\mathbf b}^{(f,i_{\textrm{un},2})} , {\mathbf b}^{(i_{\textrm{un},1},i_{\textrm{un},2})}]^T,
\end{equation}
and
$
{\mathbf h} =[{h}^{(f,i_{\textrm{un},1})} , {h}^{(f,i_{\textrm{un},2})} , {h}^{(i_{\textrm{un},1},i_{\textrm{un},2})}]^T,
$
using  \eqref{b_def_DOP+shad}, \eqref{h_def_DOP+shad}, 
 \eqref{b_def_shad+shad}, \eqref{h_def_shad+shad} to define the vector fields ${\mathbf b}$ and the scalar fields $h$ that appear in ${\mathbf B}$ and ${\mathbf h}$.

\subsection{Proposed 3: Most constrained formulation}

Our final proposed method combines all of the previous constraints, leading to a problem of  the form 
\eqref{PDE_pb_matrix}    
with
\begin{equation}\label{b_def_more_constr}
{\mathbf B} \!=\!
\begin{bmatrix}
  {b}^{(f,i_{\textrm{un},1})}_1 & {b}^{(f,i_{\textrm{un},1})}_2 \\
  {b}^{(f,i_{\textrm{un},2})}_1 & {b}^{(f,i_{\textrm{un},2})}_2 \\
  {b}^{(i_{\textrm{un},1},i_{\textrm{un},2})}_1 & {b}^{(i_{\textrm{un},1},i_{\textrm{un},2})}_2 \\
 -\cos{\phi}  & \sin{\phi} \\
\end{bmatrix},
\,\,
{\mathbf h} = 
\begin{bmatrix}
 h^{(f,i_{\textrm{un},1})} \\
  h^{(f,i_{\textrm{un},2})} \\
 h^{(i_{\textrm{un},1},i_{\textrm{un},2})} \\
    0 \\
\end{bmatrix}.
\end{equation}
This formulation uses the most information and so is potentially the most robust method. However, it requires known albedo in order to use the DOP ratio constraint. Nevertheless, it is possible to first apply proposed method 1, estimate the albedo and then re-estimate surface height using the maximally constrained formulation and the estimated albedo map. In fact, the best performance is obtained by iterating these two steps, alternately using the surface height estimate to compute albedo and then using the updated albedo to re-compute surface height.

\subsection{Extension to colour images}

We now consider how to extend the above systems of equations when colour information is available. If a surface is lit by a coloured point source, then each pixel provides three equations of the form in \eqref{eq:lamb}. In principle, this provides no more information than a grayscale observation since the surface normal and light source direction are fixed across colour channels. However, in the presence of noise using all three observations improves robustness. In particular, if the albedo value at a pixel is lower in one colour channel, the signal to noise ratio will be worse in that channel than the others. For a multicoloured object, it is impossible to choose a single colour channel that provides the best signal to noise ratio across the whole object. For this reason, we propose to use information from all colour channels where available. 

We already exploit colour information in the estimation of the polarisation image in Sec.~\ref{sec:multichanpol}. Hence, the phase angle estimates have already benefited from the improved robustness. Both the DOP ratio and intensity ratio constraints can also exploit colour information by repeating each constraint three times, once for each colour channel. In the case of the intensity ratio, the colour albedo once again cancels if ratios are taken between the same colour channels under different light source directions.

\section{Height estimation via linear least squares }\label{min_SfP_noBC}
We have seen that each of  the  variants illustrated in the previous section, each with different advantages, can be written as a PDE system \eqref{PDE_pb_matrix}. 
Denoting by $M$ the number of pixels, we discretise the gradient in \eqref{PDE_pb_matrix} via finite differences, arriving at the following linear system in $\mathbf{z}$
\begin{equation}\label{lin_min_pb}
{\mathbf A} \mathbf{z} = \bar{\mathbf{h}},
\end{equation}
where $\mathbf{A} = \bar{\mathbf{B}} \mathbf{G}$, with $\mathbf{G} \in \mathbb{R}^{2M\times M}$ the matrix of finite difference gradients.  $\bar{\mathbf{B}} \in \mathbb{R}^{JM\times 2M}$ is the discrete per-pixel version of the matrix $\mathbf{B}(\mathbf{x})$, hence $\mathbf{A} \in \mathbb{R}^{JM\times M}$, where $J$ depends on the various proposed cases reported in Sec.~\ref{sec:several_combinations} ($J=2$ for \eqref{b_def_eccv_case} and \eqref{b_def_albedo_ind}, $J=3$ for \eqref{b_def_phase_ind} and $J=4$ for \eqref{b_def_more_constr}). $\bar{\mathbf{h}}$ is the discrete per-pixel version of the function $\mathbf{h}(\mathbf{x})$, $\bar{\mathbf{h}} \in \mathbb{R}^{JM\times 1}$, and ${\mathbf z} \in \mathbb{R}^{M\times 1}$ the vector of the unknown height values. 
The resulting discrete system is large, since we have $JM$ linear equations in $M$ unknowns, but sparse, since $\mathbf{A}$ has few non-zero values for each row, and has as unknowns the height values.  
The per-pixel matrix $\mathbf{A}$ is a full-rank matrix, for each choice of $\bar{\mathbf{B}}$ that comes from the proposed formulations in Sec.~\ref{sec:several_combinations}, under the different assumptions specified for each case. 
The per-pixel matrix $\mathbf{A}$ related to \cite{SRT16} is full-rank except in one case: when the first two components of the light vector ${\mathbf s}$ are non-zero and $s_1 = -s_2$ and it happens that the phase angle is $\phi = \pi/4$ at least in one pixel. In that case, the matrix has a rank-deficiency (though in practice $\phi$ assuming a value of exactly $\pi/4$, up to numerical tolerance, is unlikely).

We want to find a solution of  \eqref{lin_min_pb} in the least-squares sense, \ie find a vector $\mathbf{z} \in \mathbb{R}^M$ such that
\begin{equation}\label{min_square}
||{\mathbf A} \mathbf{z} - \bar{\mathbf{h}}||^2_2 \leq ||{\mathbf A} \mathbf{y} - \bar{\mathbf{h}}||^2_2, \quad \forall \mathbf{y} \in \mathbb{R}^M.
\end{equation}
Considering the associated system of  normal equations
\begin{equation}\label{normal_min_system}
{\mathbf A}^T ({\mathbf A} \mathbf{z} - \bar{\mathbf{h}}) = 0,
\end{equation}
it is well-known that if there exists $\mathbf{z} \in \mathbb{R}^M$ that satisfies \eqref{normal_min_system}, then $\mathbf{z}$ is also solution of the least-squares problem, \ie $\mathbf{z}$ satisfies \eqref{min_square}. 
Since ${\mathbf A}$ is  a full-rank  matrix, then the matrix ${\mathbf A}^T {\mathbf A}$ is not singular, hence there exists a unique solution $\mathbf{z}$ of \eqref{normal_min_system} for each data term $\bar{\mathbf{h}}$. 
Since neither ${\mathbf B}$ nor ${\mathbf h}$ depend on $z$ in \eqref{PDE_pb_matrix}, the solution can be computed only up to an additive constant (which is  consistent  with the orthographic projection assumption). 
To resolve the unknown constant, 
knowledge of $z$ at just one pixel is sufficient.  
In our implementation, we remove the height of one pixel from the variables and substitute its zero value elsewhere.

\section{Two source lighting estimation}\label{sect:light_estimation}

Our three proposed shape estimation methods require knowledge of the two light source directions. Previously, Smith \etal \cite{SRT16} showed that a single polarisation image can be used to estimate illumination conditions up to a binary ambiguity. 
However, to do so, they assumed that the albedo was known or uniform,  and they also worked only with a single colour channel. In a two source setting, we show that it is possible to estimate both light source directions simultaneously, and do so in an albedo invariant manner. Moreover, we can exploit information across different colour channels to improve robustness to noise. Hence, our three methods can be used in an uncalibrated setting.

The intensity ratio \eqref{eq:shad+shad_constr} provides one equation per pixel relating unpolarised intensities, surface gradient and light source directions. Given two polarisation images with different light directions, we have one such equation per pixel and six unknowns in total.
We
assume that ambiguous surface gradient estimates are known from $\rho$ and $\phi$, and then  use \eqref{eq:shad+shad_constr} to estimate the light source directions.

\begin{figure*}[t]
\centering
\includegraphics[width=\textwidth]{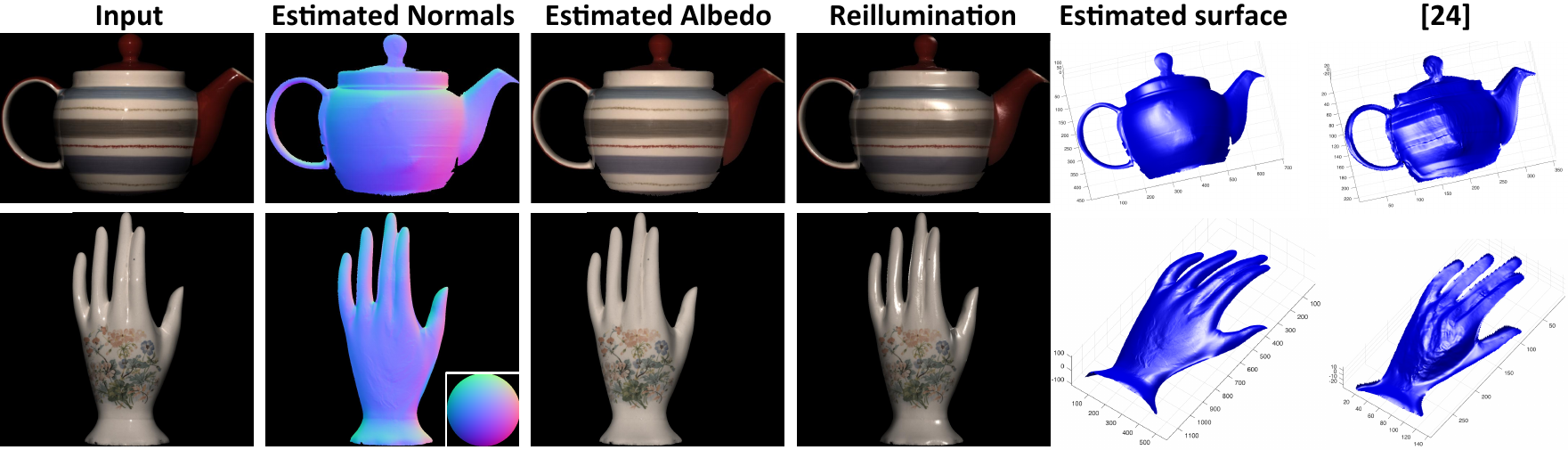}\vspace{-0.1cm}
\caption{Qualitative results on real objects with varying albedo obtained by using Prop.~1+3 and comparison to \cite{SRT16} (zoom for detail).}\vspace{-0.3cm}\label{fig:real}
\end{figure*}

The intensity ratio \eqref{eq:shad+shad_constr} is homogeneous in ${\bf s}$ and ${\bf t}$ and so has a trivial solution ${\bf s}={\bf t}=[0\ 0\ 0]^T$. If we assume that the intensity of the light source remains constant in each colour channel across the two images, then this intensity divides out when taking an intensity ratio and so the length of the light source vectors is arbitrary. We therefore constrain them to unit length (avoiding the trivial solution), and represent them by spherical coordinates $(\theta_s,\alpha_s)$ and $(\theta_t,\alpha_t)$, such that $[s_1,s_2,s_3]=[\cos\alpha_s\sin\theta_s,\sin\alpha_s\sin\theta_s,\cos\theta_s]$ and $[t_1,t_2,t_3]=[\cos\alpha_t\sin\theta_t,\sin\alpha_t\sin\theta_t,\cos\theta_t]$.
This reduces the number of unknowns to four. We can now write the residual at each pixel given an estimate of the light source directions. There are two possible residuals, depending on which of the two ambiguous polarisation normals we use. From the phase angle and the zenith angle estimated from the degree of polarisation using \eqref{eq:polariz_constraint}, we have two possible surface normal directions at each pixel and therefore two possible gradients: 
$z_x({\mathbf x}) \approx \pm \cos\phi({\mathbf x})\tan\theta({\mathbf x})\label{eqn:ambiggrad}$,   $z_y({\mathbf x}) \approx \pm \sin\phi({\mathbf x})\tan\theta({\mathbf x})$. 
Hence, the residuals at pixel ${\mathbf x}_j$ in channel $c$ are given by either:
\begin{small}
\begin{align*}
       r_{j,c}(\theta_s,\alpha_s,\theta_t,\alpha_t) = 
    &i_{\textrm{un},1}^c({\mathbf x}_j)(-z_x({\mathbf x}_j)t_1-z_y({\mathbf x}_j)t_2+t_3) -\\
&    i_{\textrm{un},2}^c({\mathbf x}_j)(-z_x({\mathbf x}_j)s_1-z_y({\mathbf x}_j)s_2+s_3),\\
&\textrm{or}\\
q_{j,c}(\theta_s,\alpha_s,\theta_t,\alpha_t) = 
    &i_{\textrm{un},1}^c({\mathbf x}_j)(z_x({\mathbf x}_j)t_1+z_y({\mathbf x}_j)t_2+t_3) -\\
&    i_{\textrm{un},2}^c({\mathbf x}_j)(z_x({\mathbf x}_j)s_1+z_y({\mathbf x}_j)s_2+s_3).
\end{align*}
\end{small}
We can now write a minimisation problem for light source direction estimation by summing the minimum of the two residuals over all pixels and colour channels:
\begin{equation*}
\min_{\theta_s,\alpha_s,\theta_t,\alpha_t} 
\sum_{j,c} \min [ r_{j,c}^2(\theta_s,\alpha_s,\theta_t,\alpha_t),q_{j,c}^2(\theta_s,\alpha_s,\theta_t,\alpha_t) ].
\end{equation*}
The minimum of two convex functions is not itself  convex and so this optimisation is non-convex. However, we find that, even with a random initialisation, it almost always converges to the global minimum. As in \cite{SRT16}, the solution is still subject to a binary ambiguity, in that if $({\bf s},{\bf t})$ is a solution then $({\bf Ts},{\bf Tt})$ is also a solution (with ${\bf T}=\textrm{diag}([-1,-1,1])$), corresponding to the convex/concave ambiguity. We resolve this simply by choosing the maximal solution when surface height is later recovered.

\section{Experiments}\label{Sec:tests}

\def\rot#1{\rotatebox{90}{#1}}
\begin{figure}
{\scriptsize {\bf
    \centering
    \begin{center}
    \begin{tabular}{c@{\hspace{0.2cm}}c@{\hspace{0.2cm}}c}
    Input & Input & Ground  \\
    (uniform albedo) & (varying albedo) & truth height \\
    \includegraphics[clip=true,height=1.8cm]{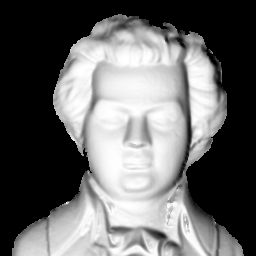} &
    \includegraphics[clip=true,height=1.8cm]{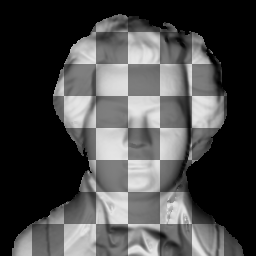} &
    \includegraphics[clip=true,height=1.8cm,trim=365px 93px 191px 150px]{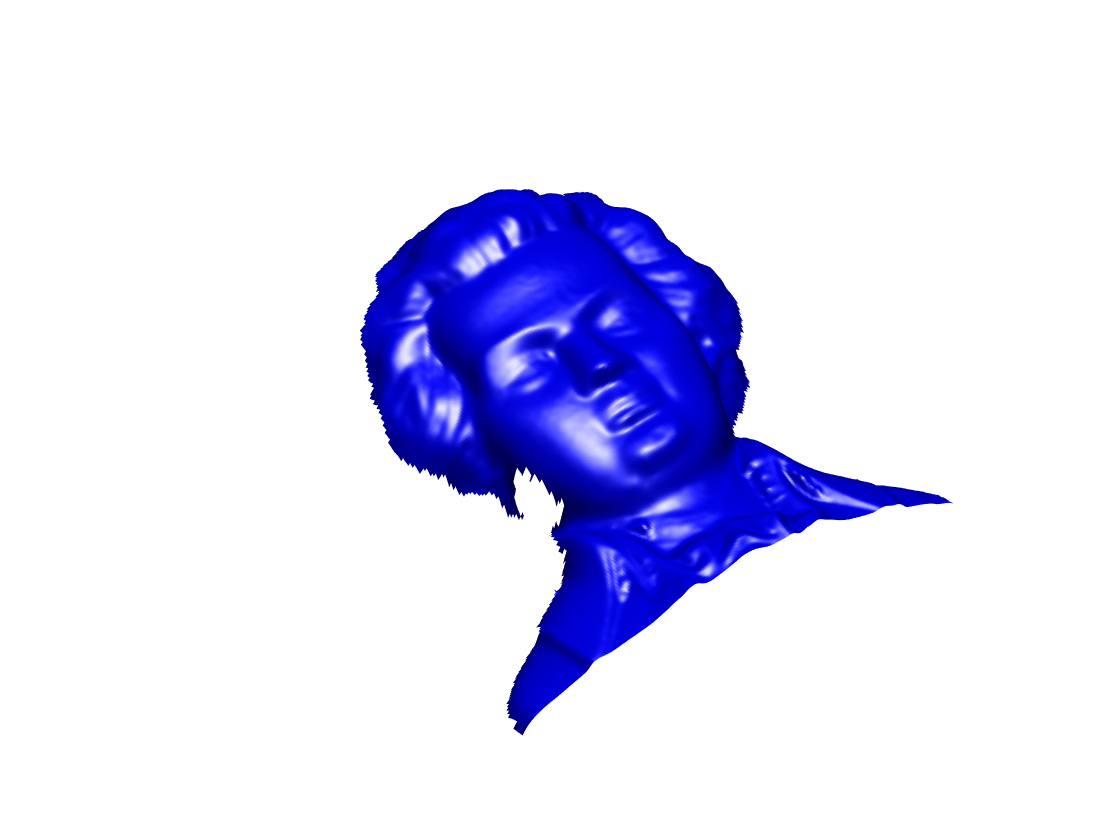}
    \end{tabular}    
    \end{center}\vspace{-0.2cm}
    \begin{tabular}{c@{\hspace{0.05cm}}c@{\hspace{0.05cm}}c@{\hspace{0.05cm}}c@{\hspace{0.05cm}}c@{\hspace{0.05cm}}c}
    & \cite{SRT16} & Prop. 1 & Prop. 2 & Prop. 3 & Prop. 1+3 \\
    \rot{Uniform albedo} &
    \includegraphics[clip=true,height=1.6cm,trim=365px 93px 191px 150px]{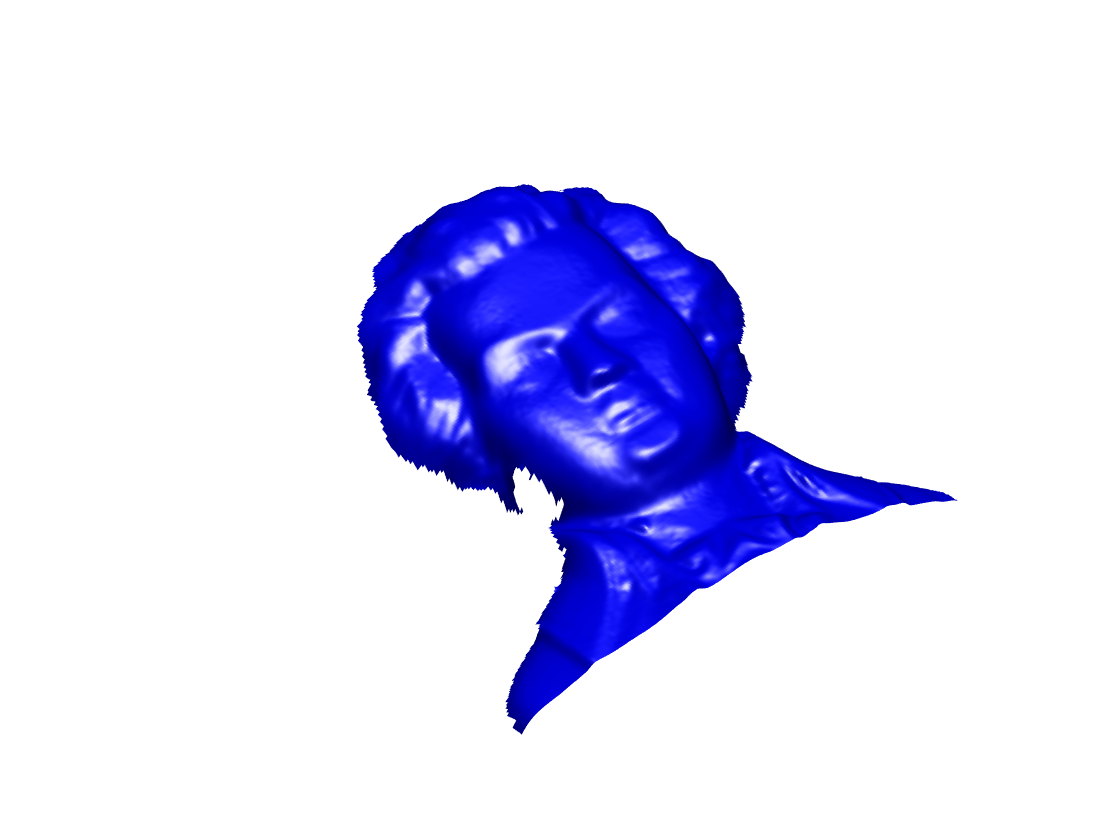} &
    \includegraphics[clip=true,height=1.6cm,trim=365px 93px 191px 150px]{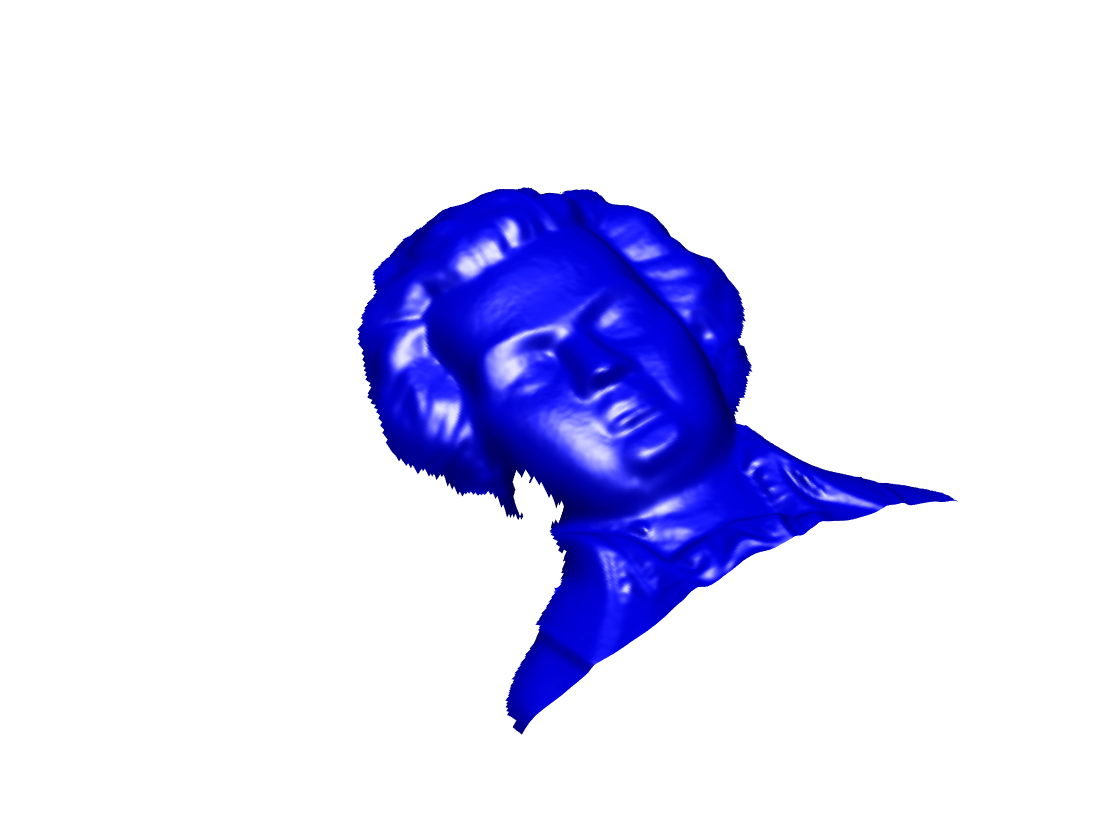} &
    \includegraphics[clip=true,height=1.6cm,trim=365px 93px 191px 150px]{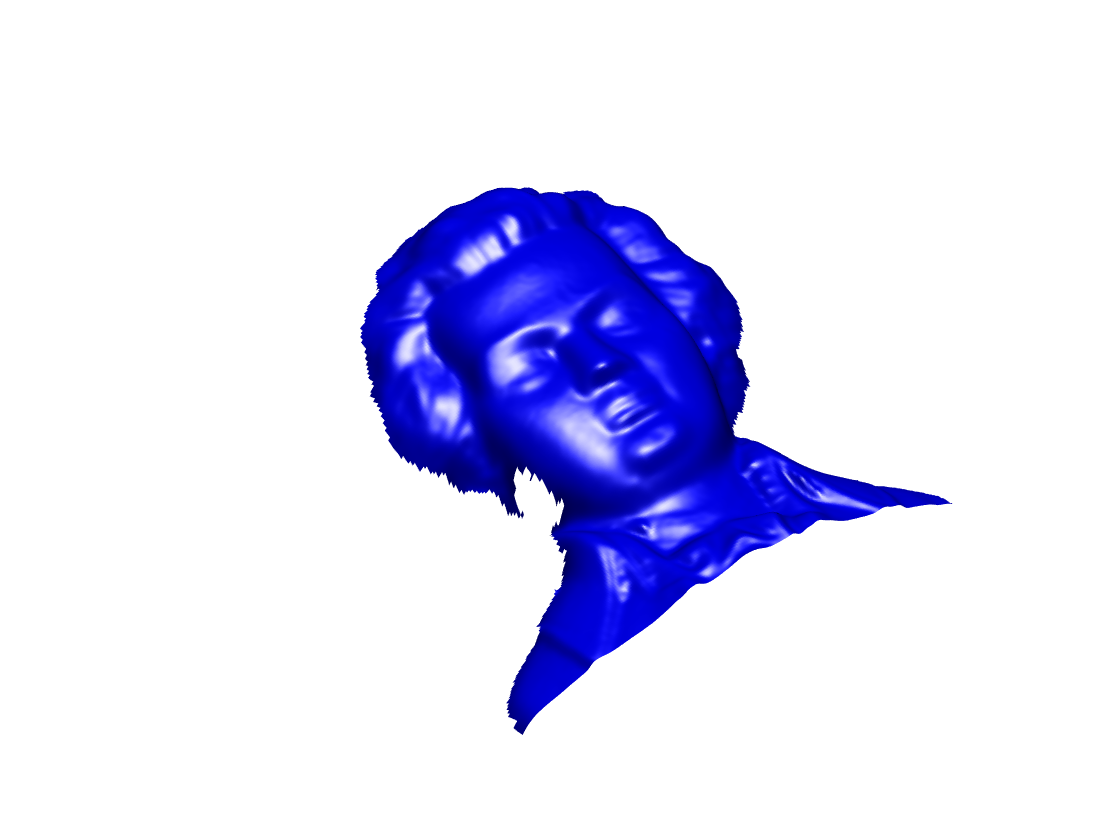} &
    \includegraphics[clip=true,height=1.6cm,trim=365px 93px 191px 150px]{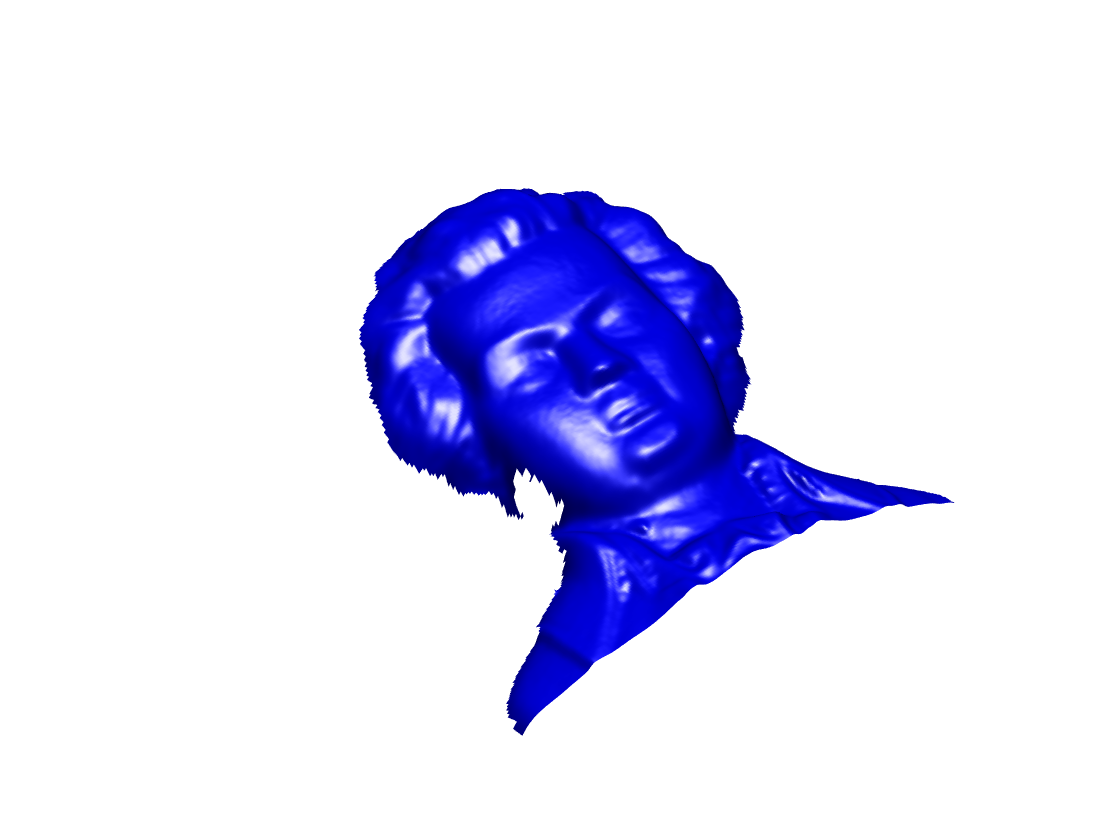} &
    \includegraphics[clip=true,height=1.6cm,trim=365px 93px 191px 150px]{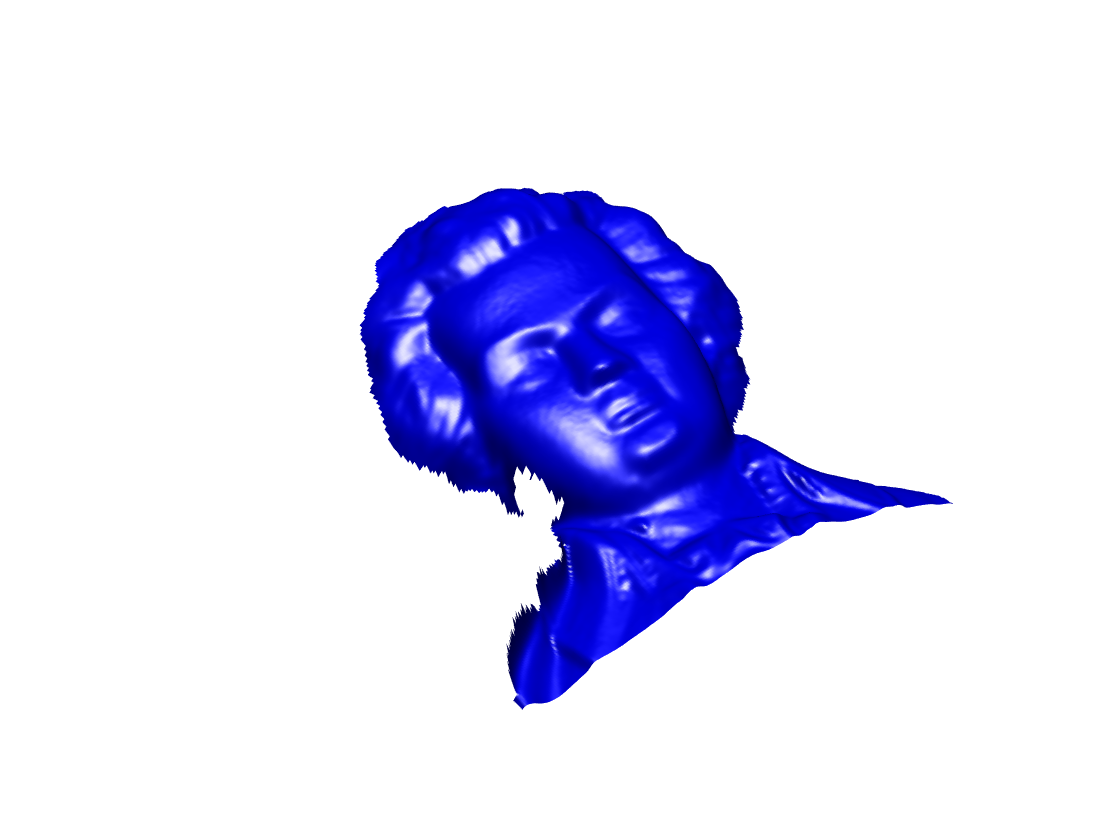} \\
    \hline
    \rot{Varying albedo} &
    \includegraphics[clip=true,height=1.6cm,trim=365px 93px 191px 150px]{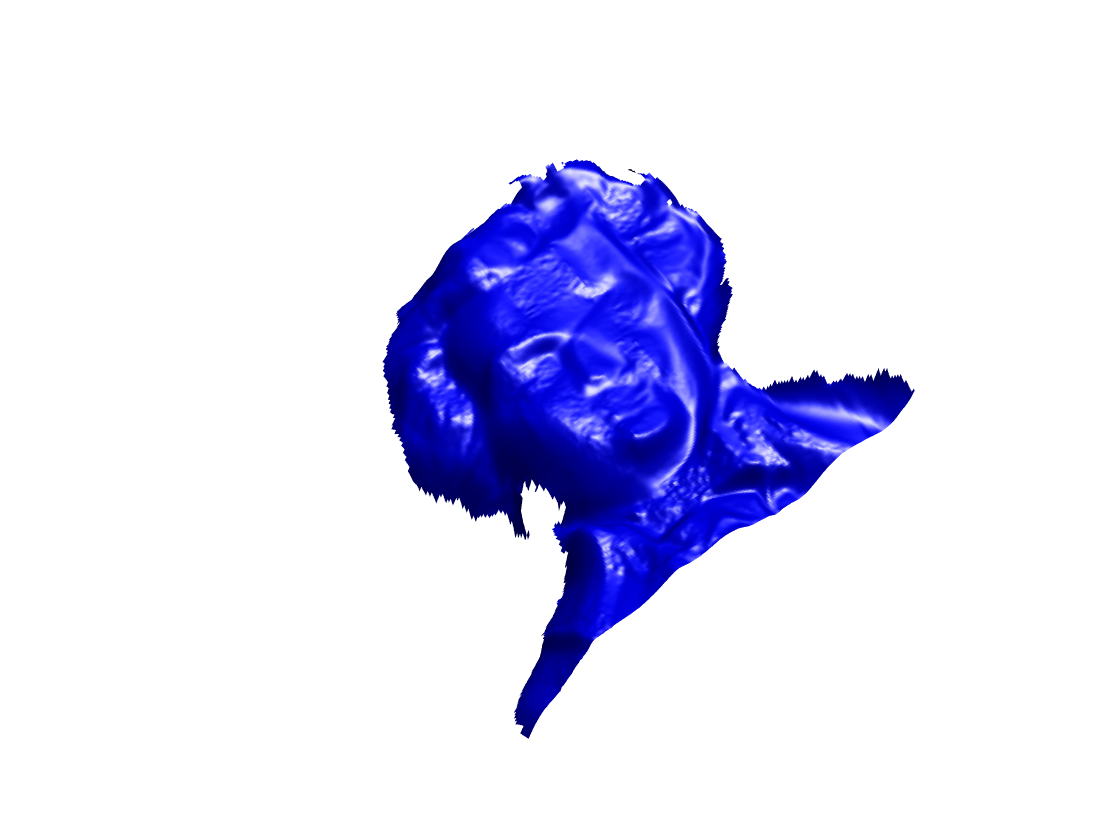} &
    \includegraphics[clip=true,height=1.6cm,trim=365px 93px 191px 150px]{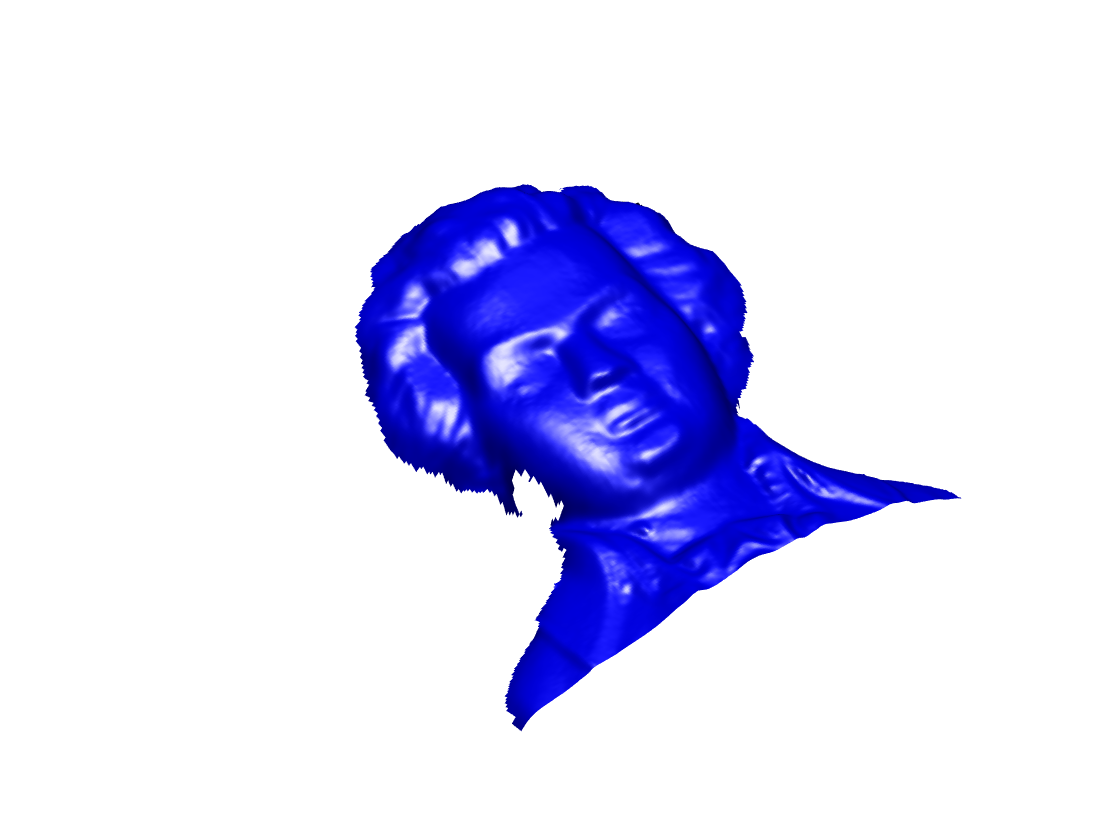} &
    \includegraphics[clip=true,height=1.6cm,trim=365px 93px 191px 150px]{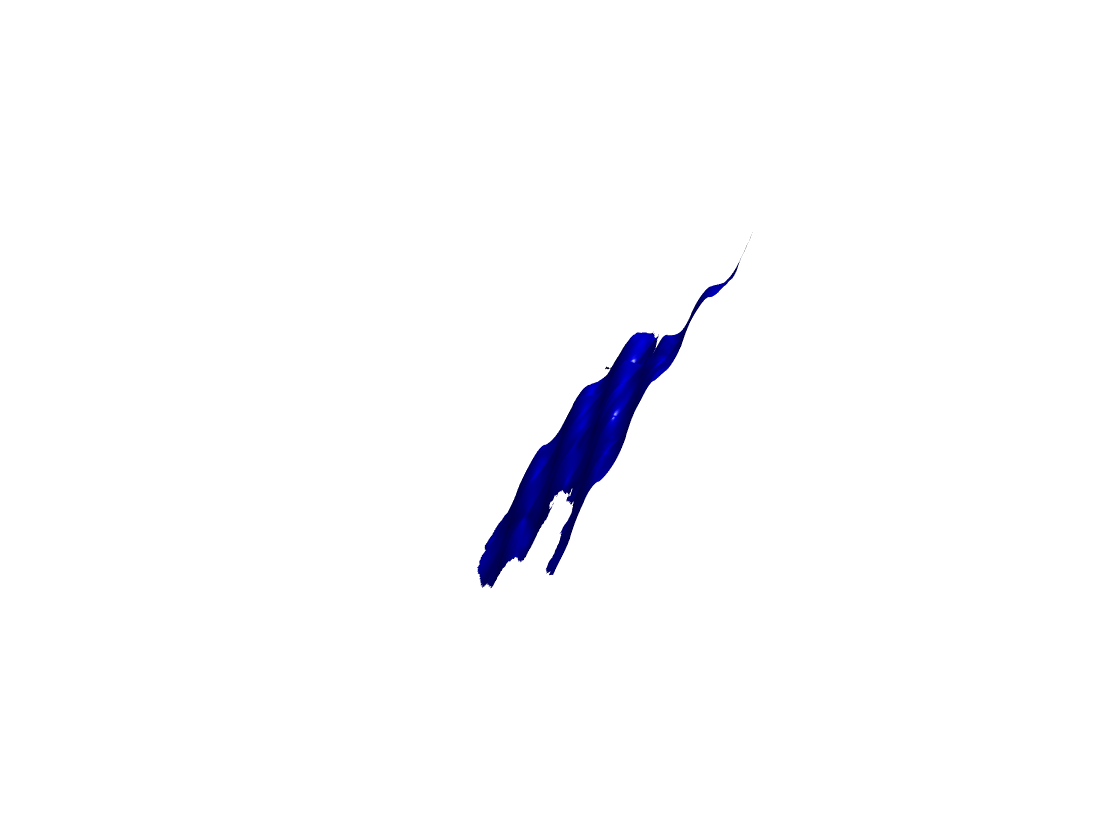} &
    \includegraphics[clip=true,height=1.6cm,trim=365px 93px 191px 150px]{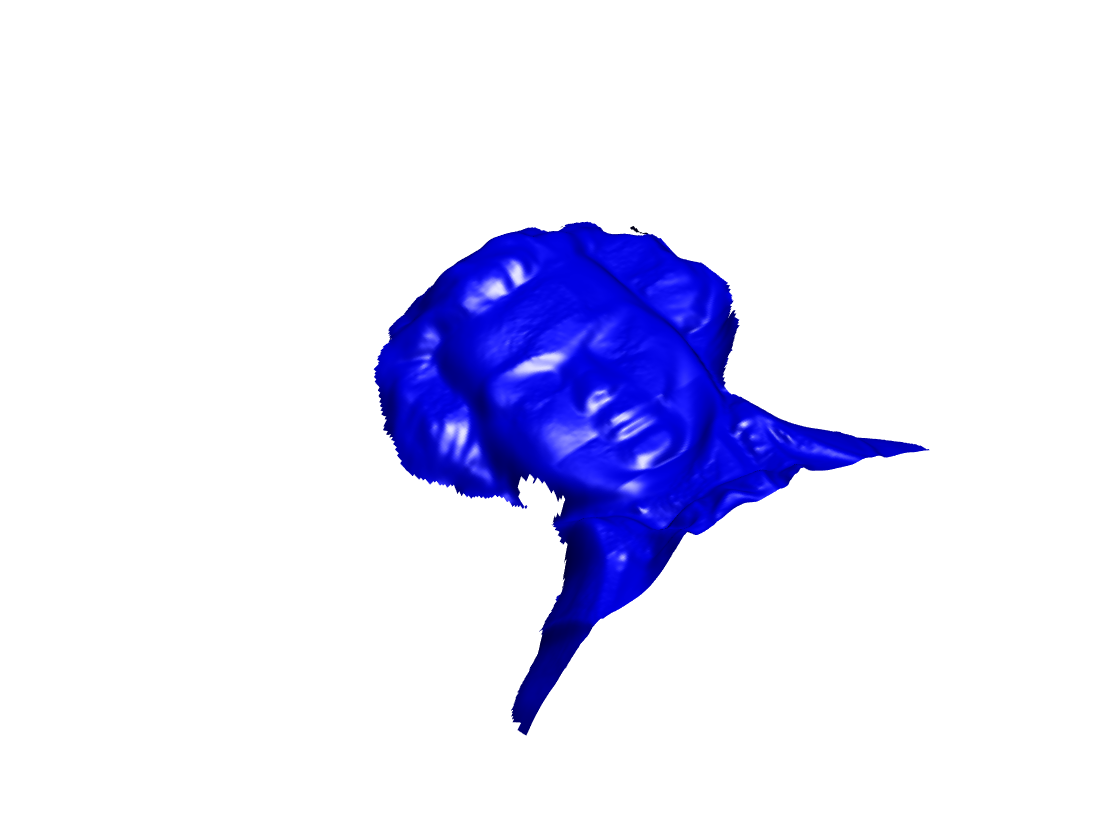} &
    \includegraphics[clip=true,height=1.6cm,trim=365px 93px 191px 150px]{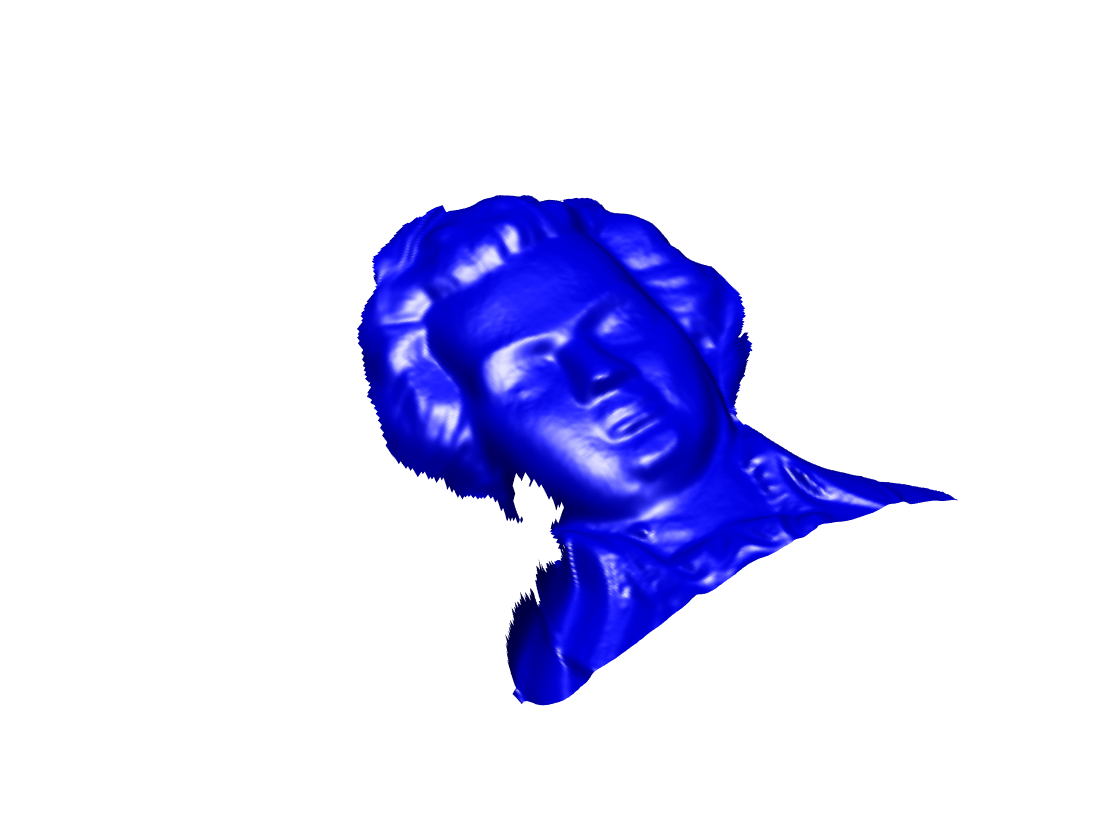} \\
    \end{tabular}
    } }\vspace{-0.2cm}
    \caption{Qualitative results on synthetic data.}\vspace{-0.2cm}
    \label{fig:synth_qual}
\end{figure}

We begin by using synthetic data generated from the Mozart height map (Fig.~\ref{fig:synth_qual}). We differentiate to obtain surface normals and compute unpolarised intensities by rendering the surface using light sources ${\bf s}=[1,0,5]^T$ and ${\bf t}=[-1,-2,7]^T$ according to \eqref{eq:lamb}. We experiment with both uniform albedo and varying albedo for which we use a checkerboard pattern. We simulate the effect of polarisation according to \eqref{eqn:TRS}, varying the polariser angle between $0^{\circ}$ and $180^{\circ}$ in $10^{\circ}$ increments. Next, we corrupt this data by adding Gaussian noise with zero mean and standard deviation $\sigma$, saturate and quantise to 8 bits. This noisy data provides the input to our reconstruction. First, we estimate a polarisation image using the method in Sec.~\ref{sec:multichanpol}, then apply each of the proposed methods or the state-of-the-art comparison method \cite{SRT16} to recover the height map. 

In Tab.~\ref{tab:quan} we report Root-Mean-Square (RMS) error in the surface height (in pixels) and mean angular error (in degrees) in the surface normals obtained by differentiating the estimated surface height. In Fig.~\ref{fig:synth_qual} we show a sample of qualitative results from this experiment. In all cases, more than one 
of our proposed methods outperform \cite{SRT16}. When albedo is uniform, our phase invariant (Prop.~2) or maximally constrained solution (Prop.~3) provides the best results. When albedo is non-uniform, the albedo invariant method (Prop.~1) provides much better performance. Although the combination of the albedo invariant method followed by the maximally constrained method (Prop.~1+3) does not give quantitatively the best performance, we find that on real world data containing more complex noise and specular reflections, this approach is most robust.

In Fig.~\ref{fig:real} we show qualitative results on two real objects with spatially varying albedo. From left to right we show: an image from the input sequence; the surface normals of the estimated height map (inset sphere shows how orientation is visualised as colour); the estimated albedo map; a re-rendering of the estimated surface and albedo map under novel lighting with Blinn-Phong reflectance \cite{Blinn77}; a rotated view of the estimated surface; and, for comparison, reconstructions of the same surfaces using \cite{SRT16}. The results of \cite{SRT16} are highly distorted in the presence of varying albedo. Our approach avoids transfer of albedo details into the recovered shape, leading to convincing relighting results.
 
\begin{table}[!t]
\centering
{\footnotesize
\setlength{\tabcolsep}{2pt}
\begin{tabular}{|c|c|c|c|c|c|c|c|}
\hline
& & \multicolumn{2}{c|}{$\sigma=0\%$} & \multicolumn{2}{c|}{$\sigma=0.5\%$} & \multicolumn{2}{c|}{$\sigma=2\%$} \\ \cline{3-8}
\multirow{2}{*}{\bf Setting} & \multirow{2}{*}{\bf Method} & Height & Normal & Height & Normal & Height & Normal \\ 
& & (pix) & (deg) & (pix) & (deg) &  (pix) & (deg) \\
\hline
\hline
\multirow{5}{*}{\parbox{1.1cm}{\centering Uniform albedo, known lighting}} 
& \cite{SRT16} & 1.12 & 2.85 & 1.68 & 4.48 & 5.06 & 11.28 \\ \cline{2-8}
& Prop. 1 & 1.78 & 2.52 & 1.94 & 3.30 & 3.49 & 7.22 \\ \cline{2-8}
& Prop. 2 & {\bf 0.23} & 1.45 & 0.70 & {\bf 1.70} & 6.50 & 5.33 \\ \cline{2-8}
& Prop. 3 & 0.42 & {\bf 1.03} & {\bf 0.52} & 1.74 & {\bf 1.53} & {\bf 4.73} \\ \cline{2-8}
& Prop. 1+3 & 3.37 & 3.22 & 3.62 & 4.03 & 5.82 & 9.15 \\ \cline{2-8}\hline
\hline
\hline
\multirow{5}{*}{\parbox{1.1cm}{\centering Uniform albedo, estimated lighting}} 
& \cite{SRT16} & 1.10 & 2.84 & 1.55 & 4.36 & 4.94 & 11.16 \\ \cline{2-8}
& Prop. 1 & 1.77 & 2.51 & 1.88 & 3.23 & 3.04 & 6.86 \\ \cline{2-8}
& Prop. 2 & {\bf 0.23} & 1.45 & 0.71 & {\bf 1.71} & 5.87 & 5.68 \\ \cline{2-8}
& Prop. 3 & 0.41 & {\bf 1.02} & {\bf 0.49} & 1.74 & {\bf 1.47} & {\bf 4.88} \\ \cline{2-8}
& Prop. 1+3 & 3.36 & 3.21 & 3.57 & 3.97 & 5.73 & 8.93 \\ \cline{2-8}
\hline
\hline
\multirow{5}{*}{\parbox{1.1cm}{\centering Unknown albedo, known lighting}} 
& \cite{SRT16} & 22.50 & 28.03 & 21.63 & 27.76 & 20.76 & 26.74 \\ \cline{2-8}
& Prop. 1 & {\bf 2.74} & {\bf 4.18} & {\bf 3.28} & {\bf 5.76} & {\bf 6.65} & {\bf 13.11} \\ \cline{2-8}
& Prop. 2 & 141.19 & 59.69 & 140.04 & 59.49 & 131.16 & 57.69 \\ \cline{2-8}
& Prop. 3 & 18.62 & 16.76 & 18.58 & 16.85 & 17.33 & 16.82 \\ \cline{2-8}
& Prop. 1+3 & 5.22 & 9.59 & 5.80 & 11.26 & 7.56 & 16.50 \\ \cline{2-8}
\hline
\hline
\multirow{5}{*}{\parbox{1.1cm}{\centering Unknown albedo, estimated lighting}} 
& \cite{SRT16} & 7.78 & 18.10 & 8.20 & 18.82 & 9.93 & 22.68 \\ \cline{2-8}
& Prop. 1 & {\bf 2.73} & {\bf 4.17} & {\bf 3.19} & {\bf 5.62} & {\bf 6.53} & {\bf 12.98} \\ \cline{2-8}
& Prop. 2 & 140.56 & 59.58 & 133.76 & 58.31 & 91.24 & 47.88 \\ \cline{2-8}
& Prop. 3 & 18.66 & 16.79 & 19.02 & 17.15 & 20.34 & 18.76 \\ \cline{2-8}
& Prop. 1+3 & 5.21 & 9.57 & 5.75 & 11.09 & 8.84 & 19.56 \\ \cline{2-8}
\hline
\end{tabular}
}
\caption{Height and surface normal errors on synthetic data.}\vspace{-0.2cm}\label{tab:quan}
\end{table}

\section{Conclusions}\label{Sec:conclusions}

In this paper we have introduced a unifying formulation for recovering height from photo-polarimetric data and proposed a variety of methods that use different combinations of linear constraints. We proposed a more robust way to estimate a polarisation image from multichannel data and showed how to estimate lighting from two source photo-polarimetric images. Together, our methods provide uncalibrated, albedo invariant shape estimation with only two light sources.  
Since our unified differential formulation does not depend on a specific camera setup or a chosen reflectance model, the most obvious target for future work is to move to a perspective projection, considering more complex reflectance models, exploiting better the information available in specular reflection and polarisation. 
In addition, since our methods directly estimate surface height, it would be straightforward to incorporate positional constraints, for example provided by binocular stereo. 

\subsubsection*{Acknowledgements}
This work was supported mainly by the ``GNCS - INdAM'', in
part by ONR grant N000141512013 and the UC San Diego Center for Visual Computing. 
W.~Smith was supported by 
EPSRC grant EP/N028481/1.

\twocolumn

{\small
\bibliographystyle{ieee}
\bibliography{biblioICCV17}
}

\end{document}